\def\year{2018}\relax
\documentclass[letterpaper]{article} 
\usepackage{aaai18}  
\usepackage{times}  
\usepackage{helvet}  
\usepackage{courier}  
\usepackage{url}  
\usepackage{graphicx}  
\usepackage{amsfonts}
\usepackage{multirow}
\usepackage{multicol}
\usepackage{amssymb}
\usepackage{amsmath}
\usepackage{makecell}
\begin{document}
%
\title{Path-Based Attention Neural Model for Fine-Grained Entity Typing}
\author{Denghui Zhang$^{1}$, Manling Li$^{1}$, Pengshan Cai$^{2}$, Yantao Jia$^{1}$, Yuanzhuo Wang$^{1}$, Xueqi Cheng$^{1}$\\
$^{1}$Institute of Computing Technology, Chinese Academy of Sciences, Beijing, 100190, China\\
$^{2}$School of Computer Science, University of Massachusetts Amherst, MA 01003\\
\{zhangdenghui, wangyuanzhuo, cxq\}@ict.ac.cn, \{limanlingcs, jamaths.h\}@gmail.com, pengshancai@cs.umass.edu\\
}
\maketitle
\begin{abstract}
Fine-grained entity typing aims to assign entity mentions in the free text with types arranged in a hierarchical structure. 
Traditional distant supervision based methods employ a structured data source as a weak supervision and do not need hand-labeled data, but they neglect the label noise in the automatically labeled training corpus. 
Although recent studies use many features to prune wrong data ahead of training, they suffer from error propagation and bring much complexity. 
In this paper, we propose an end-to-end typing model, called the path-based attention neural model (PAN), to learn a noise-robust performance by leveraging the hierarchical structure of types. Experiments 
demonstrate its effectiveness. 
\end{abstract}

\section{Introduction}
\noindent Fine-grained entity typing aims to assign types (e.g., ``person'', ``politician'', etc.) 
to entity mentions in the local context (a single sentence), and the type set constitutes a tree-structured hierarchy (i.e., type hierarchy). 
Recent years witness the boost of neural models in this task, e.g., \cite{Shimaoka2016Neural} employs an attention based LSTM to attain sentence representations and achieves state-of-the-art performance. However, it still suffers from noise in training data,  which is a main challenge in this task. 
The training data is generated by distant supervision, which assumes that if an entity has a type in knowledge bases (KBs), then all sentences containing this entity will express this type. 
This method inevitably introduces irrelevant types to the context. 
For example, the entity ``Donald Trump'' has types ``person", ``businessman'' and ``politician'' in KBs, thus all three types are annotated for its mentions in the training corpora. But in sentence ``Donald Trump announced his candidacy for President of US.'', only ``person'' and ``politician'' are correct types, 
while ``businessman'' can not be deduced from the sentence, thus serves as noise. 
To alleviate this issue, a few systems try to denoise training data by filtering irrelevant types ahead of training. 
For instance, \cite{Ren2016Label} proposes PLE to  identify correct types by jointly embedding mentions, context and type hierarchy, and then use clean data to train classifiers. 
However, the denoising and training process are not unified, which may cause error propagation and bring much additional complexity.

Motivated by this, we propose an end-to-end typing model, called the path-based attention neural model (PAN), 
to select relevant sentences to each type, which can dynamically reduce the weights of wrong labeled sentences for each type during training. 
This idea is inspired by some successful attempts to reduce noise in relation extraction, e.g.,\cite{Lin2016Neural}. 
However, these methods fail to formulate type hierarchy, which is distinct in fine-grained entity typing. 
Specifically, 
if a sentence indicates a type, its parent type can be also deduced from the sentence. 
Like the example above, ``politician'' is the subtype of ``person''. Since the sentence indicates that ``Donald Trump'' is ``politician'',  ``person'' should also be assigned. 
Thus, we build path-based attention for each type by utilizing its path to its coarsest parent type (e.g., “person”, “businessman”) in the type hierarchy. Compared to the simple attention in relation extraction, 
it enables parameter sharing for types in the same path. With the support of hierarchical information of types, 
it can reduce noise effectively and yields a better typing classifier. 
Experiments on two data sets validate effectiveness of PAN. 

\section{Path-Based Attention Neural Model}
\begin{figure}[!h]
\centering
\scalebox{0.55}{\includegraphics{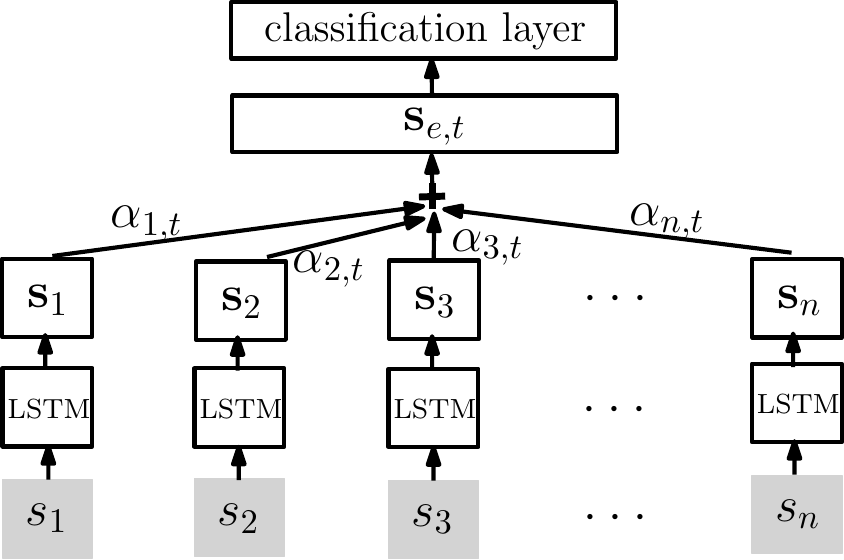}}
\vspace{-12pt}
\caption{The architecture of PAN for given entity $e$, type $t$}
\label{nn}
\vspace{-10pt}
\end{figure}

The architecture of PAN is illustrated in \figurename \ref{nn}. 
Supposing that there are $n$ sentences containing entity $e$, i.e., $\mathcal{S}_e=\{s_1,s_2,...,s_n\}$, and $\mathcal{T}_e$ is the automatically labeled types based on KBs. 
Firstly PAN employs LSTM to generate representations of sentences $\mathbf{s}_i$ following \cite{Shimaoka2016Neural}, where  $\mathbf{s}_i \in \mathbb{R}^{d}$ is the semantic representation of $s_i$, $i \in \{1,2,...,n\}$. 
Afterwards, we build path-based attention $\alpha_{i,t}$ over sentences $s_i$ for each type $t  \in \mathcal{T}_e$, which is expected to focus on relevant sentences to type $t$. 
Then, the representation of sentence set $\mathcal{S}_e$ for type $t$, denoted by $\mathbf{s}_{e,t} \in \mathbb{R}^{d}$, 
is calculated through weighted sum of vectors of sentences. 
Finally, we obtain predicted types through a classification layer. 
\vspace{-10pt}

More precisely, given $e$, an attention $\alpha_{i,t}$ is learned to score how well sentence $s_i$ matches type $t$, i.e., 
\[
\alpha_{i,t} = \frac{\exp(\mathbf{s}_i\mathbf{A}\mathbf{p}_t)}{\sum_{j=1}^{n} \exp(\mathbf{s}_j\mathbf{A}\mathbf{p}_t)},
\vspace{-4pt}
\]
where $\mathbf{A} \in \mathbb{R}^{d \times d}$ is a weighted diagonal matrix. 
$\mathbf{p}_t \in \mathbb{R}^{d}$ is the representation of path $p_t$ for type $t$. 
Specifically, for each type, we define one path as a sequence of types starting from its coarsest parent type, and ending with it. 
More formally, for type $t_l$, $p_{t_l}=t_1$$\rightarrow$$t_2$$\rightarrow$$...$$\rightarrow$$t_l$, where $t_1$ is its coarsest parent type, and $t_{i+1}$ is the subtype of $t_{i}$. 
For example, for type $t_l=politician$, its path is $p_{t_l}=person$$\rightarrow$$politician$.
We represent the path $p_{t_l}$ as a semantic composition of all the types on the path, i.e., $\mathbf{p}_{t_l}=\mathbf{t}_1 \circ \mathbf{t}_2 \circ...\circ \mathbf{t}_l$, 
where $\mathbf{t}_i \in \mathbb{R}^{d}$ is the representation of type $t_i$, which is a parameter to learn. $\circ$ is a  composition operator. 
In this paper, we consider two  types of operators: (1) Addition (PAN-A), where $\mathbf{p}_{t_l}$ equals the sum of type vectors. 
(2) Multiplication (PAN-M), where $\mathbf{p}_{t_l}$ equals the cumulative product of type vectors. 
In this way, path-based attention enables the model to share parameters between types in the same path. For example, the attention learned for ``person'' could assist the learning of the attention for ``politician''. It makes learning easier especially for infrequent subtypes, which suffer from dearth of training data, since the attentions for these subtypes can get support from the attention for parent type. 

Then, the representation of sentence set $\mathcal{S}_e$ for type $t$, i.e., $\mathbf{s}_{e,t}$, is calculated through weighted sum of sentence vectors,
\vspace{-6pt}
\[
\mathbf{s}_{e,t}=\sum_{i=1}^{n} \alpha_{i,t}\mathbf{s}_i. 
\vspace{-6pt}
\]

Since one mention can have multiple types, we employ a classification layer consisting of $N$ logistic classifiers, where $N$ is the total number of types. Each classifier outputs the probability of respective type, i.e., 
\vspace{-4pt}
\[
P(t|\mathbf{s}_{e,t})=\frac{\exp(\mathbf{w}_{t}^\mathrm{T} \mathbf{s}_{e,t}+\mathbf{b}_{t})}{1+\exp(\mathbf{w}_{t}^\mathrm{T} \mathbf{s}_{e,t}+\mathbf{b}_{t})},
\vspace{-4pt}
\]
where $\mathbf{w}_{t}, \mathbf{b}_{t}  \in \mathbb{R}^{d}$ are the logistic regression parameters. 
To optimize the model, a multi-type loss is defined according to the cross entropy as follows,
\vspace{-4pt}
\[
J=-\sum_{e} \sum_{t} [\mathbb{I}_{t}\ln P(t|\mathbf{s}_{e,t})+(1-\mathbb{I}_{t})\ln(1-P(t|\mathbf{s}_{e,t}))],
\vspace{-4pt}
\]
where $\mathbb{I}_t$ is indicator function to indicate whether $t$ is the annotated type of entity $e$, i.e., $t\in \mathcal{T}_e$. 

\section{Experiments and Conclusion}
Experiments are carried on two widely used datasets OntoNotes and FIGER(GOLD), and the training dataset of OntoNotes is noisy compared to FIGER(GOLD) \cite{Shimaoka2016Neural}. The statistics of the datasets are listed in \tablename \ref{dataset}. 
\vspace{-25pt}
\begin{table}[h!]
\small
\centering
\caption{Statistics of the datasets.}
\begin{tabular}{p{1.9cm}|p{0.75cm}p{0.75cm}p{0.83cm}p{0.78cm}p{0.78cm}}
\hline
Datasets & \#Type &\#Layer & \#Context & \#Train & \#Test  \\
\hline
OntoNotes & 89 & 3& 143K & 223K & 8,963  \\
FIGER(GOLD) &  113 & 2& 1.51M & 2.69M & 563  \\
\hline
\end{tabular}
\label{dataset}
\vspace{-14pt}
\end{table}

We employ Strict Accuracy (Acc), Loose Macro F1 (Ma-F1), and Loose Micro F1 (Mi-F1) as evaluation measures following \cite{Shimaoka2016Neural}. Specifically, ``Strict" evaluates on the type set of each entity mention, while ``Loose'' on each type. ``Marco'' is the geometric average over all mentions, while ``Micro'' is the arithmetic average. 
The baselines are chosen from two aspects: (1) Predicting types in a unified process using raw noisy data, i.e., TLSTM \cite{Shimaoka2016Neural}, and other methods shown in \tablename \ref{performance}. 
(2) Predicting types using clean data by denoising ahead, i.e., H\_PLE and F\_PLE \cite{Ren2016Label}. To prove the superiority of path-based attention, we also directly apply the attention neural model in relation extraction \cite{Lin2016Neural} without using type hierarchy (AN). 
The results of baselines are the best results reported in their papers. 
\vspace{-17pt}
\begin{table}[htbp]
\small
\centering
\caption{Performance on FIGER(GOLD) and OntoNotes}
\label{performance}
 \begin{tabular}{p{1cm}|p{0.61cm}p{0.87cm}p{0.87cm}|p{0.60cm}p{0.87cm}p{0.87cm}}
\hline
\multirow{2}{*}{Metric} & \multicolumn{3}{c|}{OntoNotes} & \multicolumn{3}{c}{FIGER(GOLD)} \\
  & Acc &  Ma-F1 & Mi-F1 & Acc & Ma-F1 & Mi-F1 \\
\hline
HYENA & 24.9 & 49.7 & 44.6 & 28.8 & 52.8 & 50.6 \\
FIGER & 36.9 & 57.8 & 51.6 &  47.4 & 69.2 & 65.5  \\
TLSTM & 50.8 & 70.1 & 64.9 &  59.7 & 79.0 & 75.4  \\
\hline
AN & 52.3 & 71.7 & 65.2 & 60.0 & 79.5 & 75.9  \\
PAN-A & \textbf{54.9} & \textbf{72.8} & \textbf{66.5} & \textbf{60.2} & \textbf{79.9} & \textbf{76.2}  \\
PAN-M &  53.0 & 71.9 &65.3 & 60.0 & 79.4 &  76.0 \\
\hline
\end{tabular}
\vspace{-10pt}
\end{table}

We can observed that: (1) When using the same raw noisy data, PAN outperforms all methods on both data sets, 
which proves the anti-noise ability of PAN. 
(2) PAN performs better than AN, since the attention learned in PAN utilizes the hierarchical structure to enable parameter sharing. 
(3) The improvements on OntoNotes are higher than FIGER(GOLD), 
because OntoNotes is more noisy, and the hierarchical structure in OntoNotes is more complex with more layers, which further demonstrates that path-based attention does well with type hierarchy, and proves the superiority of PAN in reducing noise. 
(4) PAN-A achieves better performance than PAN-M, which shows that addition operator can better capture type hierarchy.

\vspace{-17pt}
\begin{table}[htbp]
\small
\centering
\caption{Performance on FIGER(GOLD) and OntoNotes}
\label{performancedenoise}
 \begin{tabular}{p{1cm}|p{0.61cm}p{0.87cm}p{0.87cm}|p{0.60cm}p{0.87cm}p{0.87cm}}
\hline
\multirow{2}{*}{Metric} & \multicolumn{3}{c|}{OntoNotes} & \multicolumn{3}{c}{FIGER(GOLD)} \\
  & Acc &  Ma-F1 & Mi-F1 & Acc & Ma-F1 & Mi-F1 \\
\hline
H\_PLE & 54.6 & 69.2 & 62.5 & 54.3 & 69.5 & 68.1 \\
F\_PLE & \textbf{57.2} & 71.5 & 66.1& 59.9 & 76.3 & 74.9 \\
\hline
PAN-A & 54.9 & \textbf{72.8} & \textbf{66.5} & \textbf{60.2} & \textbf{79.9} & \textbf{76.2}  \\
\hline
\end{tabular}
\vspace{-10pt}
\end{table}

As shown in \tablename  \ref{performancedenoise}, 
PAN using raw noisy data outperforms H\_PLE and F\_PLE using denoised data on Ma-F1 and Mi-F1. 
It makes sense that F\_PLE has higher Acc on OntoNotes since the noise is reduced before training, but it needs to learn additional parameters about mentions, context and types, 
while PAN only needs to learn parameters of attention. Thus, PAN  is more efficient to reduce noise. 

In conclusion, PAN can reduce noise effectively through an end-to-end process, and achieves better typing performance on datasets with more noise. 



\bibliography{reference}
\bibliographystyle{aaai}

\end{document}